\documentclass[11pt,a4paper]{article}

\usepackage[hyperref]{eacl2021}
\usepackage{times}
\usepackage{latexsym}

\usepackage{url}
\usepackage{amsmath}

\usepackage{graphicx}
\usepackage{subfigure}
\usepackage{threeparttable}
\usepackage{tablefootnote}
\usepackage{url}

\graphicspath{{Fig/}}

\newcommand{\matr}[1]{\mathbf{#1}} 

\usepackage{microtype}

\aclfinalcopy 

\title{Transformer based Multilingual document Embedding model}
\author{Wei Li \textnormal{and} Brian Mak \\
Department of Computer Science and Engineering \\
The Hong Kong University of Science and Technology \\
{\tt \{wliax, mak\}@cse.ust.hk}}

\date{}

\begin{document}
\maketitle

\begin{abstract}
    One of the current state-of-the-art multilingual document
embedding model LASER is based on the bidirectional LSTM neural machine translation model. This paper presents a transformer-based sentence/document embedding model, T-LASER, which makes three significant improvements. Firstly, the BiLSTM encoder is replaced by the attention-based transformer structure, which is more capable of learning sequential patterns in longer texts. Secondly, due to the absence of recurrence, T-LASER enables faster parallel computations in the encoder to generate the text embedding. Thirdly, we augment the NMT translation loss function with an additional novel distance constraint loss.  This distance constraint loss would further bring the embeddings of parallel sentences close together in the vector space; we call the T-LASER model trained with distance constraint, cT-LASER. Our cT-LASER model significantly outperforms both BiLSTM-based LASER and the simpler transformer-based T-LASER. 
\end{abstract}

\section{Introduction}
\label{intro}




Cross-lingual text embedding model can project texts from different languages into a common vector space and produce a language-independent representation of the words \cite{klementiev2012inducing,coulmance2016trans,ap2014autoencoder,berard2016multivec} or sentences/documents\cite{hermann2014multilingual,pham2015learning,mogadala2016bilingual,berard2016multivec}.   
Such language-independent representation enables comparison, indexing, and knowledge sharing between texts of different languages \cite{hermann2014multilingual,ap2014autoencoder,mogadala2016bilingual,berard2016multivec,ferreira2016jointly,gouws2015bilbowa,artetxe2018margin,schwenk2018corpus,sinoara2019knowledge}. Cross-lingual text embedding model is especially useful when we want to transfer knowledge from resource-rich language to resource-scarce languages.

LASER is the current state-of-the-art multilingual sentence/document embedding model \cite{schwenk2017learning,artetxe2019massively};
it is designed after multilingual neural machine translation (NMT). 
It uses a bidirectional-LSTM (BiLSTM) encoder and an LSTM decoder, and the sentence embedding vector is obtained from max-pooling the encoder outputs. 
Its traits (uniform dictionary, fully shared encoder, and pivot languages) make LASER uniquely adapted to dealing with a large number of languages in multilingual situations \cite{schwenk2017learning,artetxe2019massively}.

The transformer architecture enjoys two significant advantages comparing to traditional LSTM-based architecture\cite{vaswani2017attention}.  Firstly, the absence of recurrent structure makes computational parallelism possible in a transformer encoder, as the successive computational state does not depend on the previous states. Secondly, the self-attention mechanism allows learning from a much broader context than LSTM as every state can attend to every state in the input sequence.

Inspired by LASER and the success of the transformer structure like BERT and XLNet \cite{devlin2018bert,yang2019xlnet}, this paper set two goals. This first goal is to introduce a transformer-based multilingual sentence embedding model (T-LASER) and compare its performance with the BiLSTM-based model (LASER). T-LASER will inherit LASER's advantage in dealing with a large number of languages in multilingual situations. At the same time, the new transformer-based model structure will make it better and faster than the BiLSTM-based baseline. Comparing to multilingual BERT\cite{devlin2018bert, MBERT}(mBERT), T-LASER adopts the multilingual NMT framework instead of the token masking and sentence entailment classification framework in BERT and mBERT. They also have a different model structure in the decoder part. We find that T-LASER can better learn from parallel corpus and offers better performance in our cross-lingual text classification experiment.

The second goal of the paper is to present a novel loss function that combines a distance constraint loss with the traditional translation loss for cross-lingual sentence embedding and discusses its effect on both LASER and T-LASER architecture. We would label the T-LASER model trained with this new loss function as cT-LASER, and the LASER model trained with this new loss function as cLASER. The distance constraint loss term makes two major contributions to our model. Firstly, it can significantly raise the performance in multilingual settings. Secondly, it can deal with the bilingual setting where the traditional multilingual NMT setting does not offer good performance.

Note that the most recent LASER encoder published in \cite{artetxe2019massively} is trained on a much larger corpus \footnote{It is the combination of Europarl, United Nations, OpenSubtitles2018, Global Voices, Tanzil and Tatoeba corpus, which consists of 93 input languages.}. We confine our pre-training corpus to Europarl as we want to compare our models with other models trained under similar conditions from previous works. Using a large corpus combination would increase the difficulty of separating the impacts of the model itself and the quality/quantity of the training data. With our limited computational resources, it is also very difficult for us to replicate or fine-tune our models with the 93 languages corpus combination(the training of one model would take 16 V100 GPUs for 5days). Therefore, We re-implement the current version of LASER according to \cite{artetxe2019massively} and trained it with the Europarl corpus as a baseline \footnote{ Europarl is large enough to support most popular large NMT models. The subset of five languages is sufficient to infer the quality of knowledge transfer between language pairs as many previous publications only work on 2--5 languages \cite{hermann2014multilingual,ap2014autoencoder,mogadala2016bilingual,berard2016multivec,ferreira2016jointly,gouws2015bilbowa,artetxe2018margin,schwenk2018corpus,sinoara2019knowledge,schwenk2018corpus}.}. As T-LASER and our LASER implementation are trained under the same training data and a uniformed training setting, it would make a fairer comparison to show the capability of different model structures.

MLdoc \cite{schwenk2018corpus} dataset on RCV1/RCV2 is used to evaluate the cross-lingual document embedding performance of T-LASER. MLdoc is an enhancement over the older CLDC dataset \cite{lewis2004rcv1,klementiev2012inducing}, by addressing the class imbalance problem and including more languages.  
The rationale behind the cross-lingual sentence/document embedding models is that task-unrelated parallel data like the Europarl corpus is easy to find. At the same time, multilingual data for a particular task/application is difficult to obtain. If the cross-lingual document embedding model can successfully learn how to transfer knowledge between languages, a simple classifier (e.g., SVM, MLP) trained on the English embeddings, for example, would be able to classify the German embeddings in the test set. This technology could greatly expand the user base of particular applications from one country to another without the need to obtain new task-specific data of the new language.

In this paper, both T-LASER and our LASER implementations are pre-trained only on the Europarl v7 parallel corpus; no task-related data is given to the embedding models. Some paper would label this setting as `zero-shot' \cite{artetxe2019massively}.

We do not adopt the task-unrelated pre-training and task-specific fine-tuning cross-lingual frameworks like BERT or XLNet \cite{devlin2018bert,yang2019xlnet}. Although the pre-training and fine-tuning frameworks may produce better performance when its encoder and classifier are optimized jointly in a single network, it is much easier to compare task-agnostic pre-trained embedding vectors across tasks. Moreover, the batch production of the task-agnostic pre-trained embeddings can be done offline. When needed, such pre-produced embeddings can be plugged swiftly into many different online applications.

   

\begin{figure}
\centerline{\includegraphics[scale=.6]{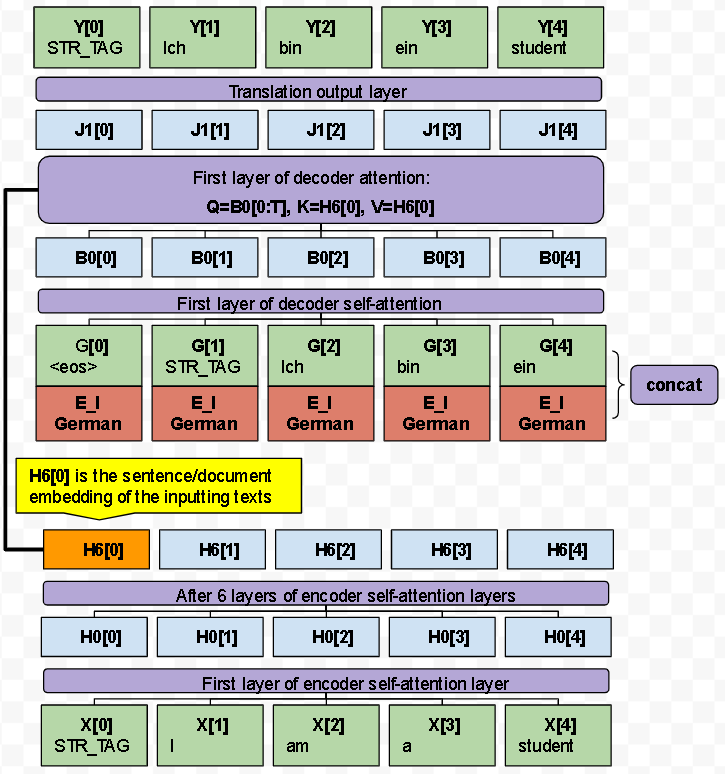}}
\caption{Architecture of T-LASER.}
\label{fig:tlaser}
\end{figure}

\section{Model Architecture}
\label{model}

Figure \ref{fig:tlaser} illustrates the architecture of T-LASER. Like LASER, T-LASER also uses shared encoder and shared decoder for all different language pairs.  A joint byte-pair encoding (BPE) vocabulary of size 50K is built from all training sentences in Europarl. No language tag/ID is provided to the encoder. As a result, it is more likely for the encoder to build a shared embedding space for all inputting languages. 

The Multi-head attention $Mh_d$ from the original transformer paper is \cite{vaswani2017attention}:
\begin{eqnarray}
h_{d_i}  =  \left(\frac{  \matr Q \matr W_i^Q (\matr K \matr W_i^K) ^T} {\sqrt{d_h}} \right) \matr V \matr W_i^V \label{eq3}  \\
Mh_d(\matr Q,\matr K,\matr V)  =   Concat(h_{d_1}, ..., h_{d_h}) \matr W^O \label{eq2}
\end{eqnarray}

where $h_{d_i}$ is $i$-th attention head, the  $ Q$, $ K$, $V$ indicate the role of  `query', `key' and `value'; $\matr  Q$, $\matr K$, $\matr V$ are matrix formed by sequentially arrange the vectors of `query', `key' and `value' ;$\matr W_i^Q$, $\matr W_i^K$, $\matr W_i^V$ are projection matrices associated with the i-th attention head; $d_h$ is the size of the encoder's hidden layers\cite{vaswani2017attention}. 

Unlike LASER, we use the first encoder output as sentence embedding and adopt a BERT-like strategy that eliminates left-padding. We insert a particular `STR\_TAG' token in the first position of every sequential text input (which can be longer than the sentence if the task is document embedding). In this way, the model would know that the first token is reserved for generating the sentence embedding. Thus the embedding of `STR\_TAG' would effectively become a learnable query fixed for every input sentence. Due to the nature of multi-head attention, each head will use this first token to query every token in the input sequence to find a weighted average of all their information from a different perspective.   
After six layers of such summarization and information extraction, the first token of the encoder out ($\matr H^6[t_0]$) would be the input text's proper sentence embedding $\vec P$. The parameters of `STR\_TAG' embedding and its associated projection matrix will learn to better summarize the overall sentence-level information during training. The self-attention and FFN in the sixth encoder layer are entirely dedicated to refining $\matr H^6[t_0]$ as we do not need to produce $\matr H^6[t_i]$ ($t_i>0$). 



Let NMT$_{a\rightarrow b}$ be the translation of a sentence from language $a$ to language $b$, the current time step be $t_i$, the target outputs be $\matr Y$.
On the decoder side, the previous target outputs $\matr G$ (where  $\matr G[t_i] = \matr Y[t_{i-1}]$ ) would first concatenate with a language embedding $\vec E_{l}$ indicating whether the current target language $b$ is English or Spanish and become $\matr J^0[t_i] $ . Without the language ID input, the decoder could generate a gibberish mixture of words from the target languages.
Then $\matr J^0[t_i]$ would first self-attend on $\matr J^0$, and the resulting vector $\matr B^k[t_i]$ would be the query of the decoder attention at $t_i$. It would attend on the $\vec P$ ($\vec P$ = $\matr H^6[t_0]$) and produce an new target output $\matr Y[t_i]$. More specifically, the $k$-th decoder layer output $\matr J^{k} $ is:
\begin{eqnarray}
\matr J^0[t_i] = Concat(\matr G[t_i],\vec E_l)
\label{eq6} \\
\matr B^k =  Mh_{d}(\matr J^{k-1},\matr J^{k-1}, \matr J^{k-1})
\label{eq9} \\
\matr C^k =  Mh_{d}(\matr  B^k,\matr H^6[t_0], \matr H^6[t_0])
\label{eq10} \\
\matr J^{k} = FFN^k(\matr C^{k})
\label{eq11}
\end{eqnarray}
where $FFN^k$ is the feed-forward network connected to the multi-head self-attention in the $k$-th layer. The $\matr J^0[t_m]$ (where ${t_m}>{t_i}$) is masked out as in an autoregressive NMT model. `\textless{eos}\textgreater' is inserted at the front of $\matr G$. 
This NMT$_{a\rightarrow b}$ translation path would result in the translation loss of $l_{mt}^{a\rightarrow b}$. Similarly , 
the translation of the same sentence from language $b$ to language $a$ (NMT$_{b\rightarrow a}$) will result in the translation loss of $l_{mt}^{b\rightarrow a}$.


As in training LASER, the source language would alternate among all the languages in training T-LASER. The target language alternates between two pivot languages, English and Spanish. For the five languages we currently use in the Europarl corpus, the training language pairs are de-en, de-es, en-es, es-en, fr-en, fr-es, it-en, and it-es \footnote{As in conventional notation, de: German, en: English, es: Spanish, Fr: French, it: Italian.}. Driven by the translation loss of the English or Spanish target of the decoder, the embeddings of every parallel input sentence (i.e., English, Spanish, French, Italian, and German translation of the same sentence) would get closer to each other \footnote {While a non-pivot language such as French would have two target languages during training, there will be only one target language (Spanish/English) for English/Spanish. }.

\subsection{Training With a Distance Constraint}




To further encourage the sentence embeddings of the same text but from different languages to get as close as possible, we propose to add a novel distance constraint loss to the training loss function. The distance loss between parallel sentence pair $d_{p}$, unrelated (negative sampled) sentence pair $d_{n_j}$ and the marginal delta loss term $\delta_{n_j}$ (the delta between $d_{p}$ and $d_{n_j}$) are computed as follows: 

\begin{eqnarray}
d_{p}= \frac{|\vec P^{(a)} - \vec P^{(b)}|^2}{v_{norm}+\epsilon} 
\label{eq41} \\
d_{n_j}^{\ a\rightarrow b}= \frac{|\vec P^{(a)}-\vec P^{(b_j)}|^2}{v_{norm}+\epsilon} 
\label{eq42} \\
\delta_{n_j}^{\ a\rightarrow b}=\max(0,\alpha - (d_{n_j}^{\ a\rightarrow b}- d_{p})) 
\label{eq43} 
\end{eqnarray}
where $\vec P^{(z)}$ is the sentence embedding from a language $z$,  $\alpha$ is the margin; $|\cdot |^2$ is the Frobenius norm of a matrix; $v_{norm}$ is the average Frobenius norm computed from all the sentences in a training batch; $\epsilon$ is a small value added to prevent division by zero; the notation $a\rightarrow b$ indicates that a quantity/variable is produced in the translation of sentences from language $a$ to language $b$.

The goal of \eqref{eq41} is to minimize the distance between the two embeddings of a related sentence pair \{$\vec P^{(a)}$, $\vec P^{(b)}$\}. 
\eqref{eq43} is used to maximize the distance between a related sentence pair and an unrelated sentence pair \{$\vec P^{(a)}$, $\vec P^{(b_j)}$\}, where $\vec P^{(b_j)}$ is the sentence embedding obtained from a randomly sampled $j$-th sentence of language $b$  \cite{hermann2014multilingual}. In \eqref{eq43},
the maximum value of  $\delta_{n_j}$  is further limited with a margin $\alpha$. To reach the marginal distance of $\alpha$ would be already good enough for distinguishing the related and unrelated sentence pairs. Further training effort could be made to bring down the translation loss and the paired distance loss $d_p$. Enhancing $\delta_{n_j}$ to the extreme will only cause the vectors' norm to become unstable.



Finally, the total loss of the model for the translation of a sentence from language $a$ to $b$, $l^{a\rightarrow b}$, 
is the sum of the paired sentence distance loss $d_{p}$, averaged $\delta_{n_j}$ from $N_s$ negative samplings and the translation cost $l_{mt}^{\ a\rightarrow b}$:
\begin{equation}
  l^{a\rightarrow b}= \beta d_{p} + \frac{\lambda}{N_s} \sum_{j=1}^{N_s} (\delta_{n_j}^{\ a\rightarrow b} + \delta_{n_j}^{\ b\rightarrow a})  +  0.5*l_{mt}^{\ a\rightarrow b}  \ ,
\label{eq44}
\end{equation}
where $0 \leq \beta \leq 1$, $0 \leq \lambda \leq 1$ are weights to balance the translation loss and the positive (paired) and negative (unrelated) distance loss. This novel cost function ensures that the produced sentence embedding vectors contain ample semantic information and are, at the same time, very similar for parallel sentences in a language pair. It is improved from the traditional distance constraint loss by further separating the roles of paired sentence distance loss $d_{p}$ and marginal delta loss $\delta_{n_j}$ and adding a new $d_{p}$ term outside of $\delta_{n_j}$. Moreover, all the loss terms are also balanced by the averaged vector norm in the batch. We find these new approaches would make the effect of the distance constraint training better and, more importantly, much more stable.



\section{Experiments}
\label{Experiment}

\begin{table*}[tbph]
	\begin{center}
		\caption{\it The dimension of the LASER and T-LASER models.}
		\begin{tabular}{|c|c|c|c|c|c|c|c|c|c|c|c|  } \hline
			Model & $d_h$ & $d_z$ & $r$ & $d_{fc}$ & \#enc & \#dec & \#vob  & \#param & WPS  & WPS$_{dist}$ 
			\\ \hline \hline
			
			  LASER$_{2018}$\shortcite{schwenk2018corpus} & 512x2 & & -  & - & 1 & 1 &  &  &   &   \\ \hline
			  LASER\_1 & 512x2 & 2048& -  & - & 1 & 1 & ~50k & 209M & 14.3K  & 12.2 K  \\ \hline
			  LASER\_6 & 512x2 & 2048& -  & - & 6 & 1 & ~50k & 241M & 12.4K  & 9.6K  \\ \hline
			  T-LASER & 1024 & 1024 & 16 & 4096 & 6 & 1 & ~50k & 246M & 23.5K & 15.2 K  \\ \hline 
		\end{tabular}
		
		\label{tbl:dimension}
	\end{center}     
\end{table*}   

We implement both T-LASER and LASER based on fairseq \cite{ott2019fairseq} \footnote{We borrow and modify the LASER training codes from \url{https://github.com/raymondhs/fairseq-laser}. The code of cT-LASER is in \url{https://github.com/ever4244/tfm_laser_0520}. }. 
Both models are trained on the Europarl v7 corpus. We take the five languages (en, de,fr, es, it), which are also in the MLdoc dataset. The same Europarl five-language subset is also used in the previous literature including LASER\cite{schwenk2018corpus, koehn2005europarl}. 

For training our LASER implementation and cLASER (LASER structure trained with distance constraint), we adopt the same settings as in \cite{artetxe2019massively} with the following additional settings due to the distance constraint loss: $N_s$ = 20, $\lambda$ = $\beta/2$, $\alpha$=0.5. While other parameters are arbitrarily decided by heuristic knowledge or related publication, $\beta$ is the main parameter for deciding the weight between the translation loss and the distance constraint loss. We sampled the models at the 10th epoch and chose a $\beta$=0.25 for cT-LASER and cLASER from $[0.25,  0.5, 1.0 ]$ according to their performances in the development set in the MLdoc dataset. For training T-LASER, we adopt the default setting as the `transformer\_vaswani\_wmt\_en\_de\_big' architecture in the fairseq\cite{ott2019fairseq,vaswani2017attention} \footnote{ adam-betas = '(0.9, 0.98)', lr (learning rate) = 0.0005, lr-scheduler = inverse\_sqrt, label-smoothing = 0.1, dropout = 0.3, weight-decay = 0.0001.}. 
Note that we applied layer normalization before each encoder/decoder block due to some NMT studies claim it is more stable in this way \cite{vaswani2017attention}. Both our LASER implementation and T-LASER have the same BPE dictionary with 50K vocabulary (Approximately 10k vocabulary for each language.) and go through the same text processing. Both LASER and T-LASER use a batch size of 128K (max number of token in a batch) and have the same uniform sampling-based data-loader and training curriculum when alternating between languages.

%
%

%
%

Table \ref{tbl:dimension} shows the model structure of the LASER and T-LASER models. LASER\_1 and LASER\_6 are implemented according to \cite{artetxe2019massively} with 1 and 6 layers of bi-directional LSTM (BiLSTM) in the encoder side respectively. $d_h$ is the size of the hidden unit of the encoder (it is also the sentence embedding size), $d_z$ is the size of the hidden unit of the decoder.  $d_{fc}$ is the dimension of FFN layers in the transformer.  Due to LASER adopts a BiLSTM encoder, which has an LSTM of size 512 for each direction, the embedding size on the encoder side is `512x2’. `\#enc,’ `\#dec,’ and `\#vob’ are the number of encoder layers, decoder layers, and vocabulary size in the model. LASER$_{2018}$ in Table \ref{tbl:dimension} is a LASER model published in \cite{schwenk2018corpus}. we use this result to examine our LASER\_1 implementation. 

Compared to the standard transformer model with six encoder and six decoder layers, our T-LASER only has one decoder layer as this model focuses on the information extraction from the encoder side. The quality of information reconstruction on the decoder side is not as important for the document embedding task as in the translation task. Albeit, T-LASER has a larger number of parameters (`\#param’). It still has faster word-per-second (WPS) and less average time per epoch (EPT) than  LASER\_6 and even LASER\_1 \footnote{ WPS is tested on one Titan RTX GPU, with CUDA 10.1, Nvidia-APEX and mix precision training (--fp16).}. This faster speed is due to T-LASER not having recurrent units as in LASER, thus better parallelism. WPS$_{dist}$ is the WPS when doing the distance constraint training \footnote{A new GPU optimized version of codes would bring the WPS gap of the  distance constraint training to less than 15\% comparing to conventional training. However, we would still report the WPS of the older version of codes here as the test performance of new codes is not fully verified yet.}.    




A new "Multilingual Document Classification Corpus" (MLdoc) dataset for cross-lingual document classification task on RCV1/RCV2 is used to evaluate the effectiveness of T-LASER \cite{schwenk2018corpus}. Compared to the old CLDC dataset \cite{lewis2004rcv1,klementiev2012inducing}, MLdoc has a uniform class distribution and supports more languages. In contrast, class `CCAT' only has less than 2\% of the examples in the old CLDC dataset. The dataset is split into 1k train set, 1k development set, and 4k test set. As in \cite{schwenk2018corpus}, we take the five testing languages which are also in the Europarl corpus. 

In a cross-lingual document classification test English-Y between English and four other languages, for example, we take a `zero-shot' test setting \cite{artetxe2019massively}. The English FFN classifier is trained with document embeddings converted from 1K English train set.  The hyperparameters are optimized on the English development set. The same English trained FFN is then attempting to classify all other input languages (German, Spanish, French, Italian). Therefore the classifier has never seen language other than English during training and hyperparameters fine-tuning \footnote{To ensure fairness; we use the same FFN and testing script from the LASER Github \url{https://github.com/facebookresearch/LASER}}. If the document embedding model can successfully project different languages into a common vector space, the FFN classifier trained on the embeddings of one language (e.g., English) would be able to classify the document embeddings of another language (e.g., German) in the test set. 


In the MLdoc experiment, we only use the first 750 tokens to produce T-LASER's document embedding in order to balance the GPU memory consumption and test accuracy. Most document in the dataset is shorter than 750 words. On the other hand, we use the full-length text for LASER\_1 and LASER\_6. Adopting a 750-word limitation would have a negative impact on LASER's result; we want our baseline results to remain strong.

\label{Result}

 \section{Results}




\begin{table}[tbph]

	\begin{center}
		\caption{\it Published MLdoc classification accuracy  (\%) of related models on the English-Y test.}
	    \vspace{-2mm}
	    
\hspace{-15mm}
\begin{tabular}{cccccc}
& de & es & fr & it & en$_{cross}$ \\
\hline
\footnotesize{MultiCCA \shortcite{ammar2016massively}} & 81.2 & 72.5 & 72.4 & 69.4 & 73.9 \\
\footnotesize{mBERT \shortcite{wu2019beto} } & 80.2 & 72.6 & 72.6 & 68.9 & 73.6 \\
\footnotesize{LASER\_5$_{2019}$\ \shortcite{artetxe2019massively} } & 84.8 & 77.3 & 78.0 & 69.4 & 77.4 \\
\footnotesize{LASER$_{2018}$* \shortcite{schwenk2018corpus}} & 71.8 & 72.8 & 66.7 & 60.7 & 68.0 \\
\footnotesize{T-LASER$_{ep13}$}* & 84.6 & 73.8 & 74.9 & 70.5 & 76.0 \\
\hline
\end{tabular}

  

		\label{tbl:publish}
	\end{center}     
\end{table}

\begin{table}[tbph]

	\begin{center}
		
		\caption{\it Published MLdoc classification accuracy (\%) of LASER$_{2018}$\shortcite{schwenk2018corpus} (trained on Europarl).}
	    \vspace{-2mm}
	    
\hspace{-10mm}	   
\begin{tabular}{ccccccc}
Train & en & de & fr & es & it & X$_{cross}$ \\
\hline
en & 88.4 & 71.8 & 72.8 & 66.7 & 60.7 & 68.0 \\
de & 71.5 & 92.0 & 75.5 & 75.5 & 56.5 & 69.7 \\
fr & 76.0 & 78.4 & 89.8 & 70.7 & 63.7 & 72.2 \\
es & 62.7 & 71.1 & 62.7 & 88.3 & 57.9 & 63.6 \\
it & 67.2 & 66.2 & 65.1 & 67.1 & 82.9 & 66.4 \\
& all: & 72.6 & same: & 88.3 & cross: & 68.0 \\
\hline
\end{tabular}

		\label{tbl:BiLSTM}
	\end{center}     
\end{table}

\begin{table*}[tbph]
	\begin{center}
		\caption{\it Comparison of MLdoc test results from T-LASER and LASER.}
	    \vspace{-2mm}


\begin{tabular}{ccccc|cccc|c}
& Dev. & & & & Best & & & & \\
\hline
model name & ep. & cross: & same: & all: & ep. & cross: & same: & all: & WPS \\
\hline
LASER$_{2018}$ &  & 68.0 & 88.3 & 72.0 & & 68.0 & 88.3 & 72.0 &  \\
LASER\_1 & 13 & 70.5 & 88.5 & 74.1 & 13 & 70.5 & 88.5 & 74.1 & 14.3K \\
LASER\_6 & 10 & 70.1 & 87.8 & 73.7 & 12 & 70.3 & 87.2 & 73.7 & 12.4K \\
cLASER\_6 & 11 & 72.6 & 87.8 & 75.6 & 11 & 72.6 & 87.8 & 75.6 & 9.6K \\
T-LASER & 13 & 73.5 & 88.6 & 76.5 & 17 & 73.9 & 88.7 & 76.9 & 23.5K \\
cT-LASER & 18 & 75.2 & 89.2 & 78.0 & 18 & 75.2 & 89.2 & 78.0 & 15.2 K \\
\end{tabular}

		\label{tbl:combine}
	\end{center}     
\end{table*}

Table \ref{tbl:publish} lists the previously published results of related works in the MLdoc experiment. The en$_{cross}$ is the average classification accuracy of the English-Y experiment, where the classifier is trained in English and tests on four other languages\footnote{ Monolingual document embedding is not this paper's focus; it does not require cross-lingual knowledge transfer. Moreover, the same language test results would be decisively affected by the vast quantity difference of pre-training data and the vocabulary size of different models. } (Y=\{de, fr, es, it\}). Table \ref{tbl:BiLSTM} presents the performance of LASER$_{2018}$ in all five language directions.  X$_{cross}$ is the average classification accuracy of the X-Y experiment, where X is the classifier's training language (see to the first column under `Train'), and Y is the four other testing languages.

LASER\_5$_{2019}$ is trained by the large 93 languages corpus combination. LASER$_{2018}$ (which has fewer encoder layers than LASER\_5$_{2019}$) and T-LASER$_{ep13}$ are trained by Europarl corpus (and are therefore marked with `*'). mBERT uses a concatenated Wikipedia corpus. Due to the huge quality and quantity difference of training data between T-LASER, mBERT, and LASER\_5$_{2019}$, it is unfair to make a direct comparison between these models.  Moreover, the  English-Y result is too biased towards English trained MLdoc performance (We find in our experiments that the overall MLdoc performance could be harmed if we tune/select the model according to English-Y). In recent studies, it is also found that choosing English as the default vector space hub of cross-lingual embedding may not be the best option and may potentially harm the performance of downstream tasks\cite{anastasopoulos2019should}.

Therefore to make a better comparison, we re-implemented the latest LASER model according to LASER\_5$_{2019}$ and trained it under the same condition, with the same Europarl corpus subset as T-LASER and LASER$_{2018}$, and listed the averaged results of all 5x5 cross-lingual test pairs in Table \ref{tbl:combine}. As a result, LASER\_6 in Table \ref{tbl:combine} and LASER\_5$_{2019}$  are essentially the same model (except that LASER\_6 has one more encoder layer). Our LASER\_6 has worse performance than LASER\_5$_{2019}$ only because Europarl is a smaller subset of the massive 93 languages corpus combination.

Table \ref{tbl:combine} summarize the MLDoc performances of the LASER and T-LASER models in one table. `ep.' is the number of epochs the model is trained. Let `Best' represents the result form the best model among 10th-20th epochs. `Dev.' represents the results from the model epochs selected by the development set. In the tables below, we reported the result under the `Dev.' setup, which would have lower performance than our `Best' models. 

`cross' is the average text classification accuracy of all cross-lingual tests (e.g., the non-diagonal elements in Table \ref{tbl:BiLSTM}; 20 in total for five languages). This result shows the ability of the model to embed text of different languages into the same vector space, thus transfer knowledge from train language to test language.  
`same' is the average text classification accuracy of five monolingual tests, in which case the FFN is trained on one language and tested on the test set of the same language. This result shows the traditional monolingual text embedding ability of the models (e.g., document vectors \cite{le2014distributed}). 
`all' is the average text classification accuracy of all 25 train-test pairs, encompassing twenty cross-lingual tests and five monolingual tests.

Comparing to LASER$_{2018}$ in Table \ref{tbl:BiLSTM}, our LASER\_1 has 2.1  better accuracy on average. This proves that our implementation and testing pipeline is successful\footnote{our LASER implementation uses two pivot languages according to LASER\_5$_{2019}$; LASER$_{2018}$ could use More. Many details are not stated in \cite{schwenk2018corpus}; a difference in batch size can also cause the performance gap .}. On the other hand, LASER\_6's performance is 1.7\% better comparing to LASER$_{2018}$. Surprisingly, LASER\_1 is 0.4\% better than LASER\_6, even though the latter has a deeper structure. After a more careful study, we find that most model epochs from LASER\_6 are still better than LASER\_1. This small performance gap could be caused by randomness in training.


Our LASER implementations are trained under a uniformed setup with T-LASER \footnote{They share the same batch size, optimizer, fairseq framework, training curriculum, random seeds, stopping criteria.}. Moreover, we also control the production of embedding during the testing stage with the same setting and processing scripts. Therefore we think it is fairer to compare the result between LASER\_1 and LASER\_6 with T-LASER and cT-LASER.

Taking consideration of the topic of this research, cross-lingual knowledge transfer, which is measured by `cross' and X$_{cross}$, are more critical performance indicators for us comparing to a simple average of `all' tests. For all 25 tests, the cross-lingual performance gap between T-LASER and LASER$_{2018}$ is about 5.5\%.  After adding our distance constraint loss term in training, cT-LASER would enjoy a more significant advantage (about 7.2\% ) in cross-lingual performance.  T-LASER and cT-LASER are better in all five train-test directions comparing to both LASER\_1 and LASER$_{2018}$. 

Although it could be argued that T-LASER has more parameters and encoder layers comparing to LASER\_1 and LASER$_{2018}$, T-LASER and cT-LASER still have a faster training speed(WPS) comparing to LASER\_1.  This is because transformer layer has better parallelism in GPU than LSTM. LASER$_{2018}$ would be even slower if it uses more than two pivot languages. 
LASER\_6 has the same number of encoder layers and similar number of model parameters comparing to T-LASER. However, its training speed (see to the WPS in Table \ref{tbl:combine}) is only half of the T-LASER.  Nevertheless, it would still be outperformed by T-LASER by 3.4\% in cross-lingual tests. The performance gap between cT-LASER (trained with distance constraint) and LASER\_6 is 5.1 \%.

Interestingly, the distance constraint training will significantly contribute to cross-lingual performance while making little or no contribution to the same language performance. The cross-lingual performance of cLASER\_6 is 2.5 percent better than the plain LASER\_6. Consider the composition of our distance constraint loss; this result is a success and reflects the goal of pulling embeddings of different languages closer. Note that the above performance gains from the distance constraint training are relatively stable to randomness in training. Even if we compare the `Dev.' performance of cT-LASER and cLASER\_6 with the `Best' performance of T-LASER and LASER\_6, the conclusion would still hold.   




\begin{table}[tbph]
	\begin{center}
		\caption{\it MLdoc classification accuracy  (\%) of cT-LASER  for 5 languages}

\hspace{10mm}	    
\begin{tabular}{ccccccc}
Train & en & de & fr & es & it & X$_{cross}$ \\
\hline
en: & 89.4 & 86.8 & 70.7 & 71.4 & 68.7 & 74.4 \\
de: & 79.0 & 93.0 & 78.4 & 76.6 & 70.0 & 76.0 \\
fr: & 78.2 & 86.9 & 88.5 & 71.7 & 64.9 & 75.4 \\
es: & 74.5 & 82.7 & 74.3 & 91.6 & 71.9 & 75.8 \\
it: & 76.0 & 79.7 & 68.8 & 73.3 & 83.5 & 74.4 \\
\hline
& all: & 78.0 & same: & 89.2 & cross: & 75.2 \\

\end{tabular}

		\label{tbl:cT-LASER}
		\vspace{-3 mm}
	\end{center}     
\end{table}

\subsection{Bilingual Experiments}
\begin{table}[tbph]

	\begin{center}
		\caption{\it Model performance trained under bilingual setting (10th epoch)}
	 
\begin{tabular}{ccc|cc}

T-LASER & & & cT-LASER & \\

Train & en & de & en & de \\
\hline
en: & 87.3 & 42.8 & 88.4 & 68.0 \\
es: & 60.6 & 91.5 & 67.7 & 91.6 \\
\hline
& same: & 89.4 & same: & 90.0 \\
& cross: & 51.7 & cross: & 67.9 \\
\end{tabular}   
        
	\label{tbl:bilan}
		\vspace{-2mm}
	\end{center}     
\end{table}   

Table \ref{tbl:bilan} shows a pair of T-LASER and cT-LASER model trained under a bilingual setting (`en-es, es-en'). This bilingual experiment would demonstrate two points. Firstly, the LASER/T-LASER framework needs pivot languages on both the target and source sides. The `en-es, es -en' training curriculum lacks the pivot language on the source side (a common input language for the same target language). Thus T-LASER has excellent same-language performance and bad cross-lingual performance. Secondly, by including the distance constraint term, we can overcome this problem and get much better cross-lingual performance (over 16\%).  In the previous five languages setting, both the translation loss and the distance constraint loss would drive the two document embedding closer. In this bilingual case, due to the lack of the pivot language on the source side,  the translation loss is mainly for the information adequacy (the same language performance). The most important driving force between cross-lingual knowledge transfer would be the distance constraint loss. Through this bilingual experiment, we can examine the effectiveness of the distance constraint training more independently. Moreover, it also shows that the distance constraint training can offer a simple solution to the cases where pivot language is absent.


\section{Conclusion}
\label{conclude}

In this paper, we presented a transformer-based document embedding model cT-LASER. It is improved from LASER and is further enhanced by a novel distance constraint loss term during training. cT-LASER is faster than the single-layer LASER\_1 and has better performance than both LASER\_1 and LASER\_6.

\clearpage




\bibliography{anthology,eacl2021}
\bibliographystyle{acl_natbib}

\appendix

\section{Appendices}
\label{sec:appendix}

\section{Supplemental Material}
\label{sec:supplemental}

\end{document}